\documentclass{article}

\usepackage[svgnames]{xcolor} 
\usepackage{graphicx}
\usepackage[margin= 0.8in]{geometry}
\usepackage{epsfig}
\usepackage{amsmath}
\usepackage{amssymb}
\usepackage{slashbox}
\usepackage{hyperref}

\newcommand*{\titleAT}{\begingroup 
\newlength{\drop} 
\drop=0.05\textheight 

\includegraphics[scale=1.5]{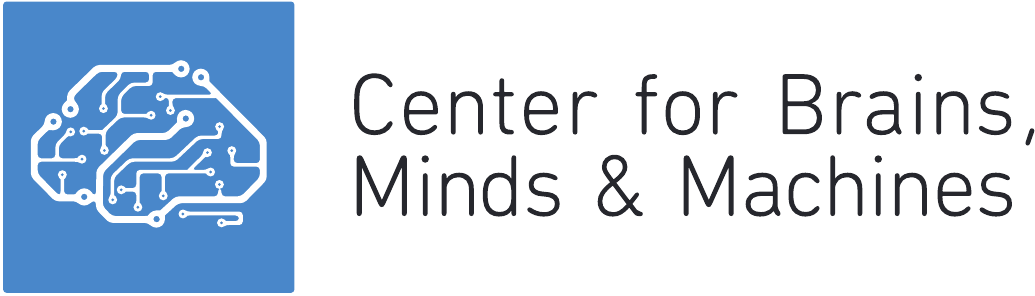}

\textcolor{CornflowerBlue}{\rule{\textwidth}{3 pt}}\par 
\vspace{2pt}\vspace{-\baselineskip} 
\rule{\textwidth}{0.4pt}\par 

\vspace{\drop} 
\textbf{\textsf{\large{CBMM Memo No. \memonumber}}}\quad \quad \quad\quad \quad \quad \quad\quad\quad \quad\quad\quad      \textbf{\large{\memodate}}

\vspace{\drop}
\begin{center}
\textbf{\textsf{\huge{\memotitle}}}\\
\vspace{0.4\drop}
\textbf{\Large{\textsf{by}}}\\
\vspace{0.4\drop}
\textbf{\textsf{\large{\memoauthors}}}
\end{center}
\vspace{\drop}
\textbf{\textsf{\large{\noindent Abstract}:}} {\memoabstract}

\textcolor{CornflowerBlue}{\rule{\textwidth}{3 pt}}\par 
\vspace{2pt}\vspace{-\baselineskip} 
\rule{\textwidth}{0.4pt}\par

\begin{minipage}{.15\linewidth}
\includegraphics[scale=0.1]{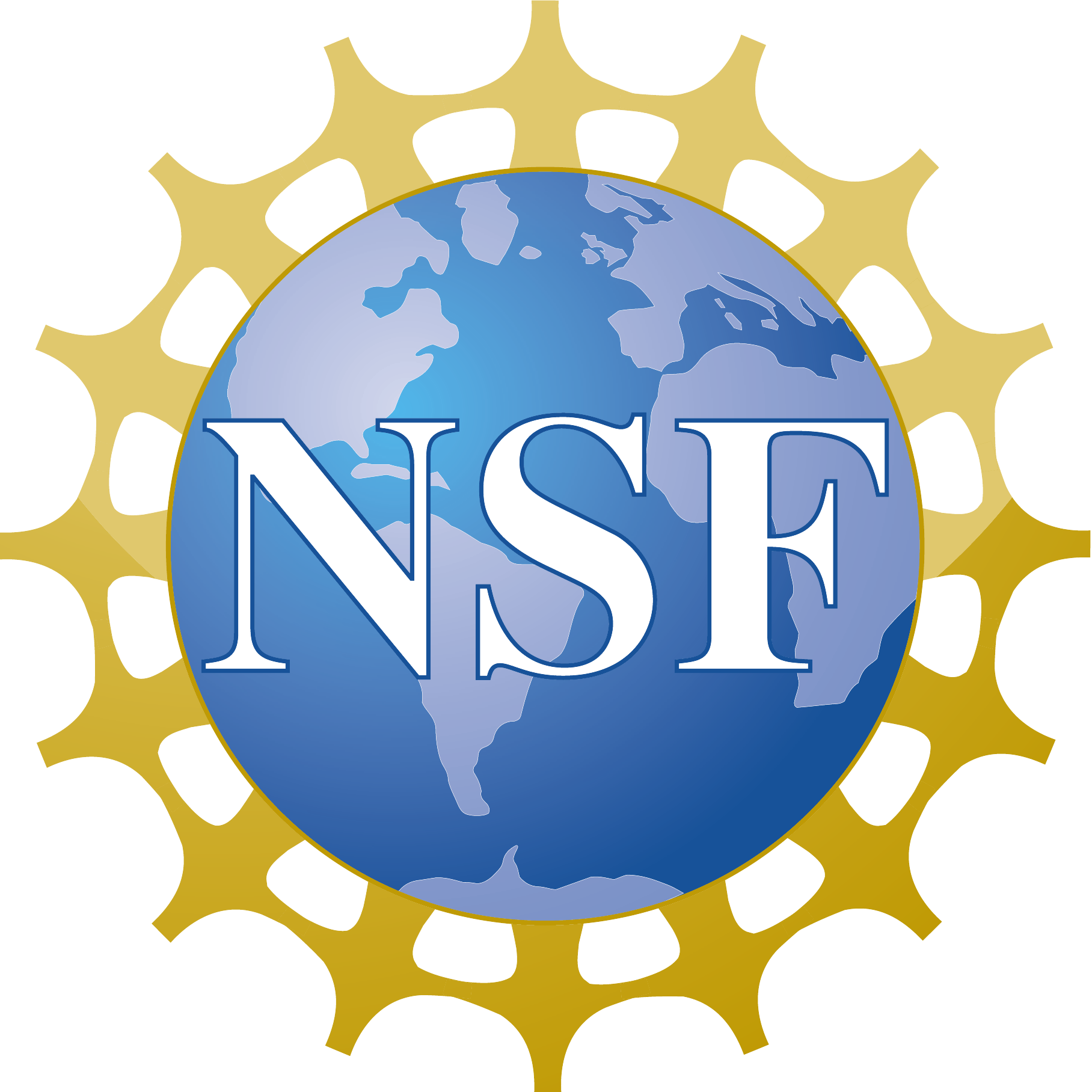}
\end{minipage}
\begin{minipage}{.84\linewidth}
\textbf{\textsf{\\
\\
The research was supported by the ONR via an award made through Johns Hopkins University, by the G. Harold and Leila Y. Mathers Charitable Foundation, by ONR N00014-12-1-0883 and the Center for Minds, Brains and Machines (CBMM), funded by NSF STC award CCF-1231216. This research was also supported by NSF Expeidtion award 1029679, ARO MURI award 58144-NS-MUR, and Intel Science and Technology Center in Pervasive Computing.}}
\end{minipage}
\endgroup}

\begin{document}

\pagestyle{empty} 

\def\eg{\textsf{\it e.g.}}
\def\ie{\textsf{\it i.e.}}

\def\memonumber{ \textsf{014}} 
\def\memodate{\textsf{\today}} 
\def\memotitle{\textsf{The Secrets of Salient Object Segmentation}} 
\def\memoauthors{\textsf{Yin Li$^{1*}$, Xiaodi Hou$^{2*}$\let\thefootnote\relax\footnote{* These authors contribute equally.}, Christof Koch$^{3}$, James M. Rehg$^{1}$, Alan L. Yuille$^{4}$ }\\
$^{1}$Gerogia Institute of Technology~~
$^{2}$California Institute of Technology \\
$^{3}$Allen Institute for Brain Science ~~
$^{4}$University of California, Los Angeles\\
{\tt\small yli440@gatech.edu~xiaodi.hou@gmail.com~christofk@alleninstitute.org~rehg@cc.gatech.edu~yuille@stat.ucla.edu}}

\def\memoabstract{\textsf{In this paper we provide an extensive evaluation of fixation prediction and salient object segmentation algorithms as well as statistics of major datasets. Our analysis identifies serious design flaws of existing salient object benchmarks, called the \emph{dataset design bias}, by over emphasising the stereotypical concepts of saliency. The dataset design bias does not only create the discomforting disconnection between fixations and salient object segmentation, but also misleads the algorithm designing.
Based on our analysis, we propose a new high quality dataset that offers both fixation and salient object segmentation ground-truth. With fixations and salient object being presented simultaneously,  we are able to bridge the gap between fixations and salient objects, and propose a novel method for salient object segmentation. Finally, we report significant benchmark progress on three existing datasets of segmenting salient objects. \\
}}

\titleAT 
\clearpage

\section{Introduction}
Bottom-up visual saliency refers to the ability to select important visual information for further processing.  The mechanism has proven to be useful for human as well as computer vision.  Unlike other topics such as object detection/recognition, saliency is not a well-defined term.  Most of the works in computer vision focus on one of the following two specific tasks of saliency: fixation prediction and salient object segmentation.

In a fixation experiment, saliency is expressed as eye gaze.  Subjects are asked to view each image for seconds while their eye fixations are recorded.  The goal of an algorithm is to compute a probabilistic map of an image to predict the actual human eye gaze patterns.  Alternatively, in a salient object segmentation dataset, image labelers annotate an image by drawing pixel-accurate silhouettes of objects that are believed to be salient.  Then the algorithm is asked to generate a map that matches the annotated salient object mask.

Various datasets of fixation and salient object segmentation have provided objective ways to analyze algorithms.  However, existing methodology suffers from two major limitations:  1) algorithms focusing on one type of saliency tend to overlook the connection to the other side.  2) Benchmarking primarily on one dataset tend to overfit the inherent bias of that dataset.

In this paper, we explore the connection between fixation prediction and salient object segmentation by augmenting 850 existing images from PASCAL 2010 \cite{pascal-voc-2010} dataset with eye fixations, and salient object segmentation labeling.  In Sec.~\ref{sec:dataset} we argue that by making the image acquisition and image annotation \emph{independent} to each other, we can avoid \emph{dataset design bias} -- a specific type of bias that is caused by experimenters' unnatural selection of dataset images.

With fixations and salient object labels simultaneously presented in the same set of images, we report a series of interesting findings.  First, we show that salient object segmentation is a valid problem because of the high consistency among labelers.  Second, unlike fixation datasets, the most widely used salient object segmentation dataset is heavily biased.  As a result, all top performing algorithms for salient object segmentation have poor generalization power when they are tested on more realistic images. Finally, we demonstrate that there exists a strong correlation between fixations and salient objects.

Inspired by these discoveries, in Sec.~\ref{sec:model} we propose a new model of salient object segmentation. By combining existing fixation-based saliency models with segmentation techniques, our model bridges the gap between fixation prediction and salient object segmentation.  Despite its simplicity, this model significantly outperforms state-of-the-arts salient object segmentation algorithms on all 3 salient object datasets.

\section{Related Works}\label{sec:relatedWorks}
In this section, we briefly discuss existing models of fixation prediction and salient object segmentation. We also discuss the relationship of salient object to generic object segmentation such as CPMC \cite{carreira2010constrained, li2010object}.  Finally, we review relevant research pertaining to dataset bias.

\subsection{Fixation prediction}
The problem of fixation based bottom-up saliency is first introduced to computer vision community by \cite{itti1998model}.  The goal of this type of models is to compute a ``saliency map'' that simulates the eye movement behaviors of human.  Patch-based \cite{itti1998model, bruce2005saliency, harel2006graph, hou2008dynamic, zhang2008sun} or pixel-based \cite{hou2012image, garcia2012relationship} features are often used in these models, followed by a local or global interaction step that re-weight or re-normalize features saliency values.

To quantitatively evaluate the performance of different fixation algorithms, ROC Area Under the Curve (AUC) is often used to compare a saliency map against human eye fixations. One of the first systematic datasets in fixation prediction was introduced in \cite{bruce2005saliency}.  In this paper, Bruce \emph{et al.} recorded eye fixation data from $21$ subjects on $120$ natural images.  In a more recent paper \cite{judd2009learning}, Judd \emph{et al.} introduced a much larger dataset with $1003$ images and $15$ subjects.

Due to the nature of eye tracking experiments, the error of recorded fixation locations can go up to $1^\circ$, or over $30$ pixels in a typical setting.  Therefore, there is no need to generate a pixel-accurate saliency map to match human data.  In fact, as pointed out in \cite{hou2012image}, blurring a saliency map can often increase its AUC score.

\subsection{Salient object segmentation}
It is not an easy task to directly use the blurry saliency map from a fixation prediction algorithm.  As an alternative, Liu \emph{et al.} \cite{liu2007learning} proposed the MSRA-5000 dataset with bounding boxes on the salient objects. Following the idea of ``object-based'' saliency, Achanta \emph{et al.} \cite{achanta2009frequency} further labeled $1000$ images from MSRA-5000 with pixel-accurate object silhouette masks.  Their paper showed that existing fixation algorithms perform poorly if benchmarked F-measures of PR curve.  Inspired by this new dataset, a line of papers has proposed \cite{cheng2011global, perazzi2012saliency, margolin2013makes} to tackle this new challenge of predicting full-resolution masks of salient objects.  An overview of the characteristics and performances of salient object algorithms can be found in a recent review \cite{borji2012salient} by Borji \emph{et al.}

Despite the deep connections between the problems of fixation prediction and object segmentation, there is a discomforting isolation between major computational models of the two types.  Salient object segmentation algorithms have developed a set of techniques that have little overlapping with fixation prediction models.  This is mainly due to a series of differences in the ground-truth and evaluation procedures.  A typical fixation ground-truth contains several fixation dots, while a salient object ground-truth usually have one or several positive regions composed of thousands of pixels.  Having different priors of sparsity significantly limited the model of one type to have good performance on tasks of the other type.

\subsection{Objectness, object proposal, and foreground segments}\label{sec:reviewObjectness}
In the field of object recognition, researchers are interested in finding objects independent of their classes~\cite{BingObj2014}.  Alexe \emph{et al.}~\cite{alexe2012measuring} used a combination of low/mid level image cues to measure the ``objectness'' of a bounding box.  Other models, such as CPMC \cite{carreira2010constrained,li2010object} and Object Proposal \cite{endres2010category}, generate segmentations of candidate objects without relying on category specific information.  The obtained ``foreground segments'', or ``object proposals'' are then ranked or scored to give a rough estimate of the objects in the scene.

The role of a scoring/ranking function in the aforementioned literature shares a lot of similarities with the notion of saliency.  In fact, \cite{alexe2012measuring} used saliency maps as a main feature for predicting objectness.  In Sec.~\ref{sec:model}, we propose a model based on the foreground segmentations generated by CPMC.  One fundamental difference between these methods to visual saliency is that an object detector is often exhaustive -- it looks for \emph{all} objects in the image irrespective of their saliency value.  In comparison, a salient object detector aims at enumerating a subset of objects that exceed certain saliency threshold.  As we will discuss in Sec.~\ref{sec:model}, an object model, such as CPMC offers a ranking of its candidate foreground proposals.  However, the top ranked (e.g. first 200) segments do not always correspond to salient objects or their parts.

\subsection{Datasets and dataset bias}\label{sec:reviewBias}
Recently, researchers started to quantitatively analyze dataset bias and their detrimental effect in benchmarking.  Dataset bias arises from the selection of images \cite{torralba2011unbiased}, as well as the annotation process \cite{tatler2005visual}.  In the field of visual saliency analysis, the most significant bias is \emph{center bias}.  It refers to the tendency that subjects look more often at the center of the screen \cite{tatler2005visual}.  This phenomenon might be partly due to experimental constraints that a subject's head being on a chin-rest during the fixation experiment, and partly due to the photographer's preference to align objects at the center of the photos.

Center bias has been shown to have a significant influence on benchmark scores \cite{judd2009learning, zhang2008sun}. Fixation models either use it explicitly \cite{judd2009learning}; or implicitly by padding the borders of a saliency map \cite{harel2006graph, hou2008dynamic, zhang2008sun}.  To make a fair evaluation of the algorithm's true prediction power, Tatler \cite{tatler2005visual} proposed a shuffled-AUC (s-AUC) score to normalize the effect of center-bias.  In s-AUC, positive samples are taken from the fixations of the test image, whereas the negative samples are from all fixations across all other images.

\section{Dataset Analysis}\label{sec:dataset}
In this paper, we will benchmark on the following datasets: Bruce \cite{bruce2005saliency}, Judd \cite{judd2009learning}, Cerf \cite{cerf2008predicting}, FT \cite{achanta2009frequency}, and IS \cite{li2013visual}.  Among these 5 datasets, Judd and Cerf only provides fixation ground-truth.  FT only provides salient object ground-truth.  IS provides both fixations\footnote{IS provides raw gaze data at every time point.  We use the following thresholds to determine a stable fixation: min fixation duration: $160 ms$, min saccade speed: $50px/100ms$.} as well as salient object masks.  While Bruce dataset was originally designed for fixation prediction, it was recently augmented by \cite{borji2013stands} with $70$ subjects under the instruction to \emph{label the single most salient object in the image}.
In our comparison experiment, we include the following fixation prediction algorithms: ITTI \cite{itti1998model}, AIM \cite{bruce2005saliency}, GBVS \cite{achanta2009frequency}, DVA \cite{hou2008dynamic}, SUN \cite{zhang2008sun}, SIG \cite{hou2012image}, AWS \cite{garcia2012relationship}; and following salient object segmentation algorithms: FT \cite{achanta2009frequency}, GC \cite{cheng2011global}, SF \cite{perazzi2012saliency}, and PCAS \cite{margolin2013makes}. These algorithms are top-performing ones in major benchmarks~\cite{borji2012salient}.

\begin{figure}[t]
\centering
\includegraphics[width=0.7\linewidth]{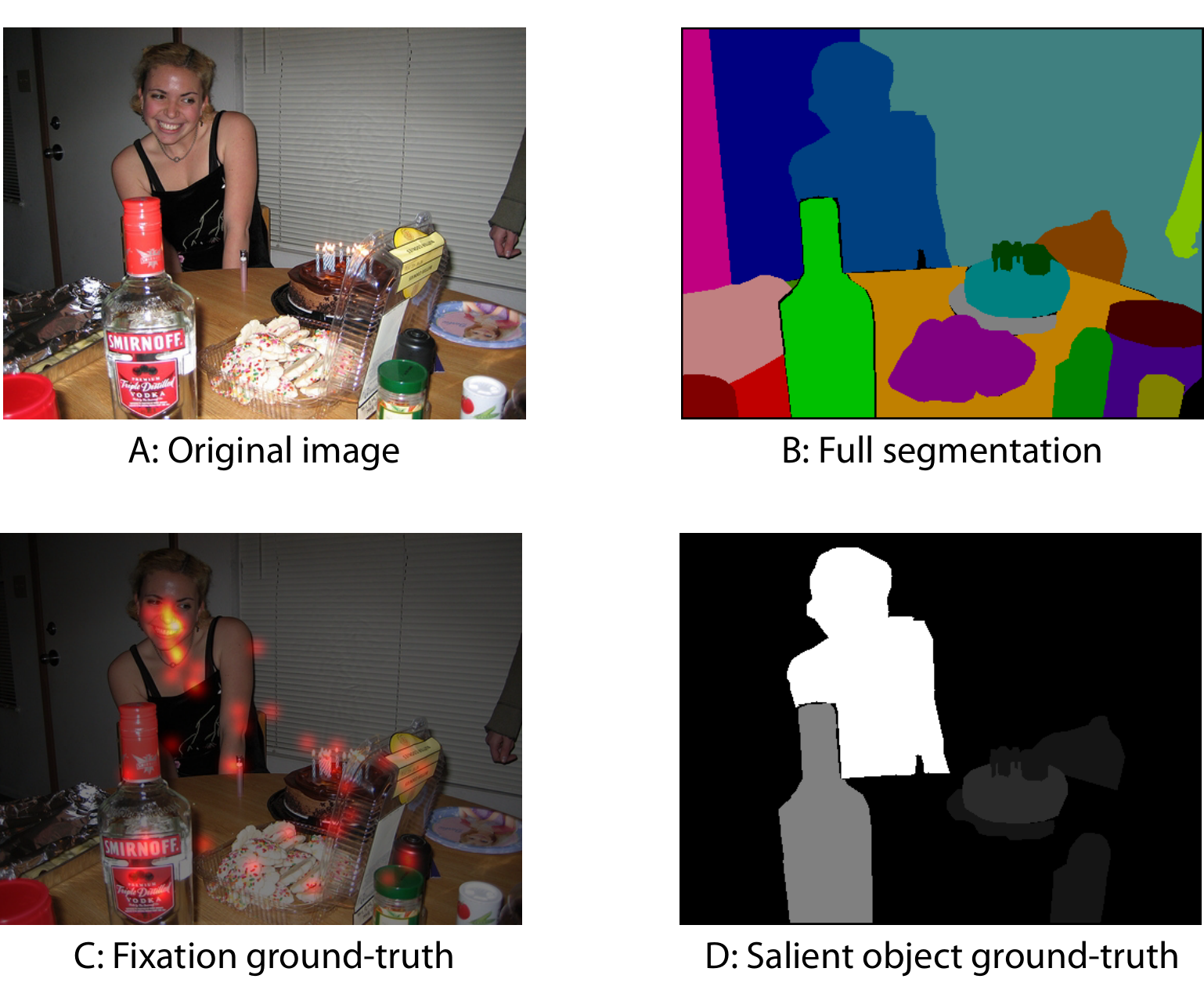}\\
\caption{An illustration of PASCAL-S dataset.  Our dataset provides both eye fixation (Fig.~C) and salient object (Fig.~D) mask.  The labeling of salient objects is based on the full segmentation (Fig.~B).  A notable difference between PASCAL-S and its predecessors is that each image in PASCAL-S is labeled by multiple labelers without restrictions on the number of salient objects.}\label{fig:objLabelGT}
\end{figure}

\subsection{Psychophysical experiments on the PASCAL-S dataset}
Our PASCAL-S dataset is built on the validation set of the PASCAL VOC 2010 \cite{pascal-voc-2010} segmentation challenge.  This subset contains $850$ natural images.  In the fixation experiment, $8$ subjects were instructed to perform the ``free-viewing'' task to explore the images.  Each image was presented for $2$ seconds, and eye-tracking re-calibration was performed on every 25 images.  The eye gaze data was sampled using Eyelink 1000 eye-tracker, at $125Hz$.  In the salient object segmentation experiment, we first manually perform a full segmentation to crop out all objects in the image.  An example segmentation is shown in Fig.~\ref{fig:objLabelGT}.B.  When we build the ground-truth of full segmentation, we adhere to the following rules: 1) we do not intentionally label parts of the image (e.g. faces of a person); 2) disconnected regions of the same object are labeled separately; 3) we use solid regions to approximate hollow objects, such as bike wheels.

We then conduct the experiment of $12$ subjects to label the salient objects.  Given an image, a subject is asked to select the salient objects by clicking on them.  There is no time limitation or constraints on the number of objects one can choose.  Similar to our fixation experiment, the instruction of labeling salient objects is intentionally kept vague.  The final saliency value of each segment is the total number of click it receives, divided by the number of subjects.

\subsection{Evaluating dataset consistency}\label{sec:consistency}
Quite surprisingly, many of today's widely used salient object segmentation datasets do not have any guarantee on the inter-subject consistency.  To compare the level of agreement among different labelers in our PASCAL-S dataset and other existing dataset, we randomly select $50\%$ of the subjects as the test subset.  Then we benchmark the saliency maps of this test subset by taking the rest subjects as the new ground-truth subset. For fixation task, the test saliency map for each image is obtained by first plotting all the fixation points from the test subset, and then filter the saliency map by a 2D Gaussian kernel with $\sigma = 0.05$ of the image width. For salient object segmentation task, the test/ground-truth saliency maps are binary maps obtained by first averaging the individual segmentations from the test/ground-truth subset, and then threshold with $Th=0.5$\footnote{At least half of the subjects within the subset agree on the mask.} to generate the binary masks for each subset.  Then we compute either AUC score or F-measure of the test subset and use this number to indicate the inter-subject consistency.

We notice that the segmentation maps of Bruce dataset are significantly sparser than maps in PASCAL-S or IS. Over $30\%$ of the segmentation maps in Bruce dataset are completely empty.  This is likely a result of the labeling process.  In Borji \emph{et al.}'s experiment \cite{borji2013stands}, the labelers are forced to choose only one object for each image.  Images with two or more equally salient objects are very likely to become empty after thresholding.  Although Bruce is one of the very few datasets that offer both fixations and salient object masks, it is not suitable for our analysis.

For dataset PASCAL-S and IS, we benchmark the F-measure of the test subset segmentation maps by the ground-truth masks.  The result is shown in Tab.~\ref{tab:consistency}.

\begin{table}[h]\centering
\begin{tabular}{c|c|c|c|c}
\hline
\hline
\multicolumn{5}{c}{AUC scores}\\
\hline
PASCAL-S & Bruce & Cerf & IS & Judd \\
\hline
0.835 & 0.830 & 0.903 & 0.836 & 0.867\\
\hline
\hline
\end{tabular}

\begin{tabular}{c}
\\
\end{tabular}

\begin{tabular}{c|c}
\hline
\hline
\multicolumn{2}{c}{F-measures}\\
\hline
PASCAL-S & IS \\
\hline
0.972 & 0.900 \\
\hline
\hline
\end{tabular}
\caption{Inter-subject consistency of 5 fixation datasets (AUC scores) and 2 salient object segmentation datasets (F-measures).}\label{tab:consistency}
\end{table}


Similar to our consistency analysis of salient object dataset, we evaluate the consistency of eye fixations among subjects (Tab.~\ref{tab:consistency}).  Even though the notion of ``saliency'' under a context of complex natural scene is often considered as ill-defined, we observe highly consistent behaviors among human labelers in both eye-fixation and salient object segmentation tasks.

\subsection{Benchmarking}\label{sec:benchmarking}
\begin{figure*}[bth]
\centering
\includegraphics[width=1.0\linewidth]{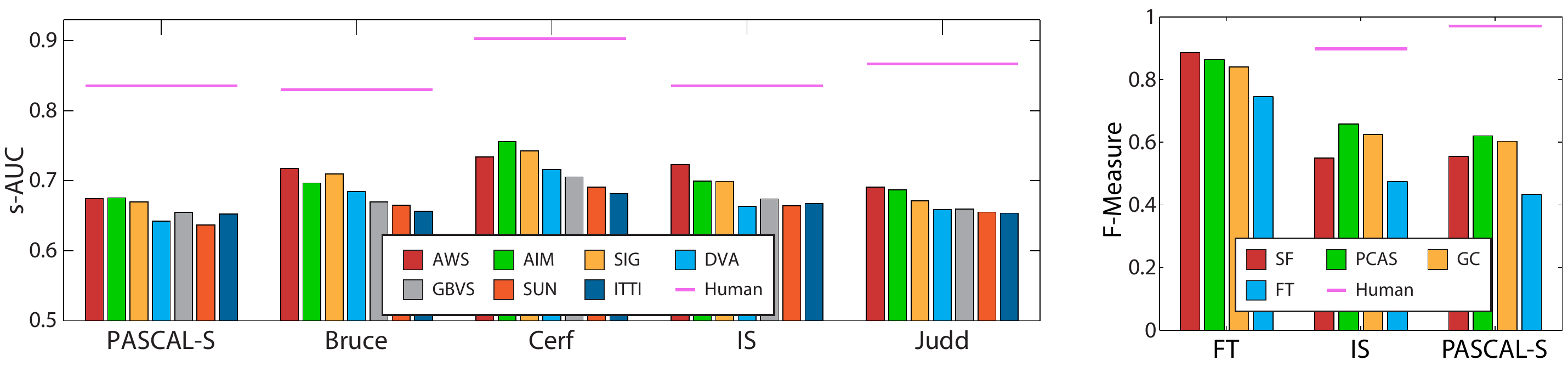}\\
\caption{\textbf{Left}: The s-AUC scores of fixation prediction.  \textbf{Right}: The F-Measure scores of salient object segmentation.  According to \cite{achanta2009frequency},  we choose $\beta = 0.3$ to calculate the F-measure from PR curve. In both figures, magenta lines show the inter-subject consistency score of these datasets. These numbers can be interpreted as the upper-bounds of algorithm scores. }\label{fig:fullBench}
\end{figure*}

In this section, we benchmark 7 fixations algorithms, AWS \cite{garcia2012relationship}, AIM \cite{bruce2005saliency}, SIG \cite{hou2012image}, DVA \cite{hou2008dynamic}, GBVS \cite{harel2006graph}, SUN \cite{zhang2008sun}, and ITTI \cite{itti1998model} on 5 datasets, Bruce \cite{bruce2005saliency}, Cerf \cite{cerf2008predicting}, IS \cite{li2013visual}, Judd \cite{judd2009learning}, and our PASCAL-S.  For salient object segmentation, we bench 4 algorithms SF \cite{perazzi2012saliency}, PCAS \cite{margolin2013makes}, GC \cite{cheng2011global}, and FT \cite{achanta2009frequency} on 4 datasets, FT \cite{achanta2009frequency}, IS \cite{li2013visual}, and our PASCAL-S.  For all algorithms, we use the original implementations from the authors' websites.  The purposes of this analysis are: 1) to highlight the generalization power of algorithms, and 2) to investigate inter-dataset difference among these independently constructed datasets.  The benchmark results are presented in Fig.~\ref{fig:fullBench}.  Sharply contrasted to the fixation benchmarks, the performance of all salient object segmentation algorithms drop significantly when migrating from the popular FT dataset.  The \emph{average} performance of all 4 algorithms have dropped, from FT's $0.8341$, to $0.5765$ ($30.88\%$ drop) on IS, and $0.5530$ ($33.70\%$ drop) on PASCAL-S.  This result is alarming, because the magnitude of the performance drop from FT to any dataset by any algorithm, can easily dwarf the 4-year progress of salient object segmentation on the widely used FT dataset.  Moreover, the relative ranking among algorithms also changes from one dataset to another.

\subsection{Dataset design bias}
The performance gap among datasets clearly suggests new challenges in salient object segmentation.  However, it is more important to pinpoint the cause to the performance degradation rather than to start benchmark race on another new dataset.  In this section, we analyze the following image statistics in order to find the similarities and differences of today's salient object segmentation datasets:

\begin{figure}[t]
\centering
\includegraphics[width=0.7\linewidth]{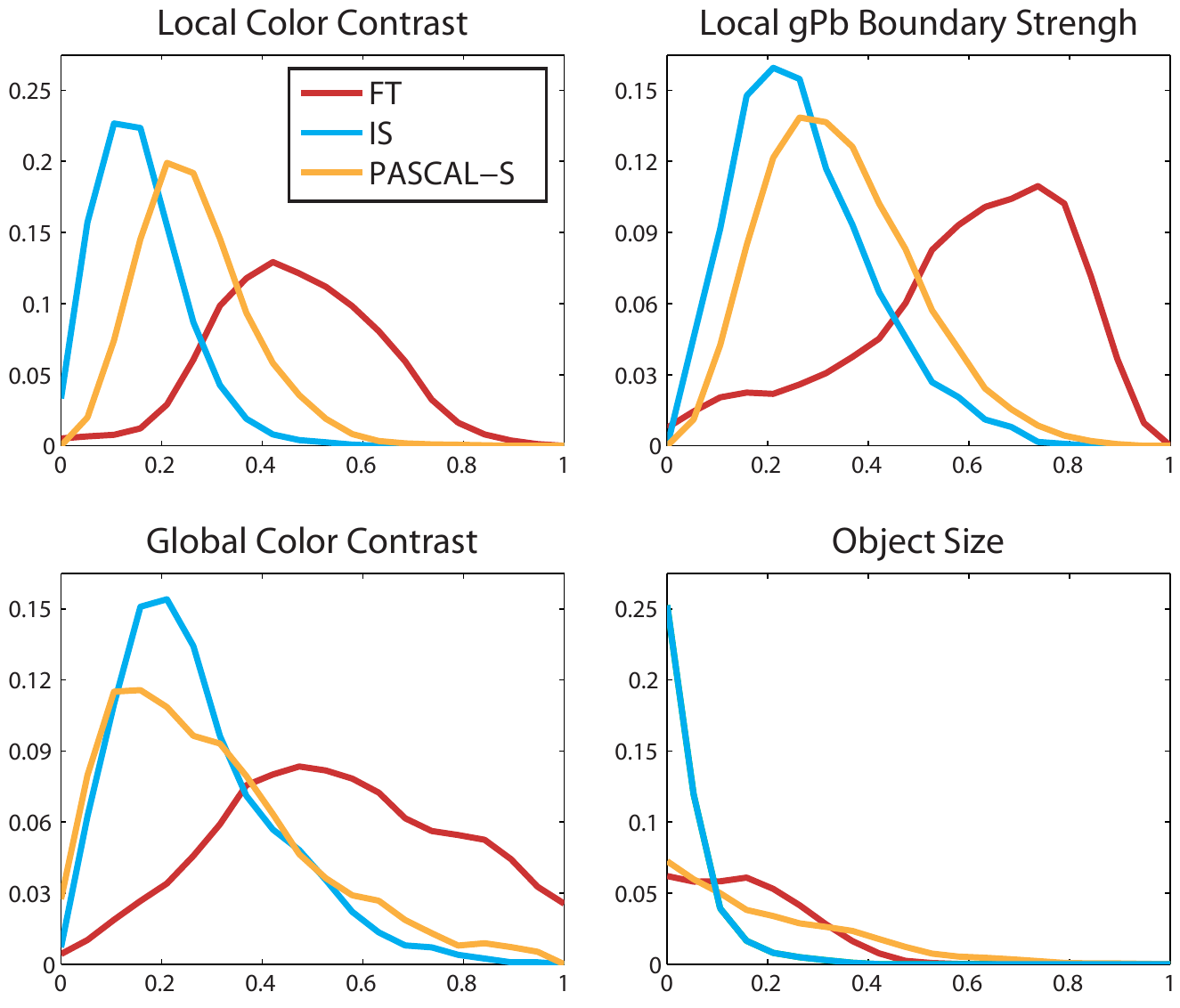}\\
\caption{Image statistics of the salient object segmentation datasets.  The statistics of FT dataset is different from other datasets in local/global color contrast and boundary strength.  As for object size, the PASCAL-S contains a balanced mix of large and small objects.}\label{fig:designBias}
\end{figure}

\begin{description}
\item [Local color contrast:] Segmentation or boundary detection is an inevitable step in most salient objects detectors.  It is important to check whether the boundaries are ``unnaturally'' easy to segment.  To estimate the strength of the boundary, we crop a $5 \times 5$ image patch at the boundary location of each labeled object, and compute RGB color histograms for foreground and background separately.  We then calculate the $\chi^2$ distance to measure the distance between two histograms.  It is worth noting that some of the ground-truth of the IS dataset are not perfectly aligned with objects boundaries, resulting in an underestimate of local contrast magnitude.
\item [Global color contrast:] The term ``saliency'' is also related to the global contrast of the foreground and background.  Similar to the local color contrast measure, for each object, we calculate the $\chi^2$ distance between its RGB histogram and the background RGB histogram.
\item [Local gPB boundary strength:] While color histogram distance captures some low-level image features, an advanced boundary detector, such as gPB \cite{arbelaez2011contour}, combines local and global cues to give a more comprehensive estimate of the presence of a boundary.  As a complementary result to our local color contrast measure, for object boundary pixel we compute the mean gPB response of a $3 \times 3$ local patch.
\item [Object size:] In each image, we define the size of an object as the proportion of pixels in the image.  Most of the objects in IS are very small.
\end{description}

As shown in Fig.~\ref{fig:designBias}, FT dataset stands out in local/global color contrast as well as the gPB boundary strength statistics.  At a first glance, our observation that FT contains unnaturally strong object boundaries seem acceptable, especially for a dataset focusing on salient object analysis.  Strong boundaries are linked to the core concept of saliency: a foreground object with discernable boundaries being surrounded by background that have contrastive colors.  In fact, many images in the FT dataset are textbook examples to demonstrate the definition of saliency.  The notion of ``saliency'' in FT is much less ambiguous than in other datasets.  However, such reduction of ambiguity during the \emph{image selection process} is more destructive rather than constructive, for the purpose to test saliency.  The confusion between image selection process and image annotation process introduces a special kind of bias by over-expressing the desired properties of the target concept, and reduce the presence of negative examples.  We call this type of bias the \emph{dataset design bias}:
\begin{quote}
During the design of a dataset, the image annotation process should be independent of the image selection process.  Otherwise the \emph{dataset design bias} will arise as a result of disproportionate sampling of positive/negative examples.
\end{quote}

\subsection{Fixations and F-measure}
Previous methods of salient object segmentation have reported big margin of F-measures over all fixation algorithms.  However, most of these comparisons are done on the FT dataset, which has shown to have non-negligible dataset bias.  Another factor that contributes to the inferior performance of fixation algorithms is the center-bias.  Major salient object segmentation algorithms, such as SF, PCAS, and GC, have implemented their own treatments for the center-bias.  In contrast, many fixation prediction algorithms, such as AWS and SIG, do not implement center bias as they expect to be benched by s-AUC score which cancels the center-bias effect.  To discount the influence of center-bias, we add a fixed Gaussian ($\sigma = 0.4$ of the image width) to the saliency maps generated by all fixation algorithms, and then benchmark these algorithms on all 3 salient object datasets.  The result is shown in Fig.~\ref{fig:finalResult}.

In addition, we also tested the F-measure of the ground-truth human fixation maps on IS and PASCAL-S.  Each fixation map is blurred by a Gaussian kernel with $\sigma = 0.03$ of the image width.  No center bias is superimposed because the human fixations are already heavily biased towards the center of the image.

When we remove the effect of center bias and dataset design bias, the performance of fixation algorithms becomes very competitive.  We also notice that the F-measure of the ground-truth human fixation maps is rather low compared to the inter-subject consistency scores in the salient object labeling experiment. This performance gap could be either due to a weak correlation between the fixation task and the salient object labeling task, or the incompatibility of the representation of fixations (dots) versus the representation of objects (regions).  In the next section, we will show that the latter is more likely to be true.  Once equipped with appropriate underlying representation, the human fixation map, as well as their algorithm approximations, generate accurate results for salient object segmentation.

\section{From Fixations to Salient Object Detection}\label{sec:model}
Many of today's well-known salient object algorithms have the following two components: 1) a suitable representation for salient object segmentation, and 2) computational principles of feature saliency, such as region contrast \cite{cheng2011global} or element uniqueness \cite{perazzi2012saliency}.  However, none of these two components alone is new to computer vision.  On one hand, detecting boundaries of objects has been a highly desired goal for segmentation algorithms since the beginning of computer vision  On the other hand, defining rules of saliency has been studied in fixation analysis for decades.  In this section, we build a salient object segmentation model by combining existing techniques of segmentation and fixation based saliency. The core idea is to first generate a set of object candidates, and then use the fixation algorithm to rank different regions based on their saliency. This simple combination results in a novel salient object segmentation method that outperforms \emph{all} previous methods by a large margin.

\subsection{Salient object, object proposal and fixations}
Our first step is to generate the segmentation of object candidates by a generic object proposal method. We use CPMC \cite{carreira2010constrained} to obtain the initial segmentations. CPMC is an unsupervised framework to generate and rank plausible hypotheses of object candidates without category specific knowledge. This method initializes foreground seeds uniformly over the image and solves a set of min-cut problems with different parameters. The output is a pool of object candidates as overlapping figure-ground segments, together with their ``objectness'' scores. The goal of CPMC is to produce a over-complete coverage of potential objects, which could be further used for tasks such as object recognition.

The representation of CPMC-like object proposal is easily adapted to salient object segmentation.  If all salient objects can be found from the pool of object candidates, we can reduce the problem of salient object detection to a much easier problem of salient segment ranking. Ranking the segments also simplifies the post-processing step. As the segments already preserve the boundary of the image, no explicit segmentation (e.g.\ GraphCut~\cite{cheng2011global}) is required to obtain the final binary object mask.

To estimate the saliency of a candidate segment, we utilize the spatial distribution of fixations within the object.  It is well known that the density of fixation directly reveals the saliency of the segment.  The non-uniform spatial distribution of fixations on the object also offers useful cues to determine the saliency of an object. For example, fixation at the center of a segment will increase its probability of being an object.  To keep our frame work simple, we do not consider class specific or subject specific fixation patterns in our model.

\subsection{The model}
We use a learning based framework for the segment selection process. This is achieved by learning a scoring function for each object candidate. Given a proposed object candidate mask and its fixation map, this function estimates the overlapping score (intersection over union) of the region with respect to the ground-truth, similar to~\cite{li2010object}.

\textbf{Features:} We extract two types of features: shape features and fixation distribution features within the object. The shape features characterizes the binary mask of the segment, which includes major axis length, eccentricity, minor axis length, and the Euler number. For fixation distribution features, we first align the major axis of each object candidate, and then extract a $4\times3$ histogram of fixations density over the aligned object mask. This histogram captures the spatial distribution of the fixations within a object.

Without loss of generality, we use the term {\it fixation energy} to refer to pixel values in the fixation map. For human fixations, the energy is a discrete value as the number of fixations at current location. For algorithm generated fixation maps, the energy is a continuous value ranging from $0$ to $1$ as the probability of a fixation at current location. Finally, the $33$ dimensional feature vector is extracted for each object mask. The details of the features can be found in Table.~\ref{tab:feat}.  In particular, the {\it Fixation Energy Ratio} is the defined as the sum of fixation energy within the segment, divided by the sum of fixation energy of the whole image.

\begin{table}
\begin{center}
\begin{tabular} {| l | c || l | c |}
\hline
\textbf{Shape Features} & Dims & \textbf{Fixation Features} & Dims\\
\hline
Area & 1 & Min/Max Fixation Energy & 2\\
\hline
Centroid &  2 & Mean Fixation Energy & 1\\
\hline
Convex Area & 1 & Weighted Fixation Centroid & 2\\
\hline
Euler Number & 1 & Fixation Energy Ratio & 1\\
\hline
Perimeter & 1 & Histogram of Fixations & 12 \\
\hline
Major/Minor Axes Length & 2 & &\\
\hline
Eccentricity & 1 & &\\
\hline
Orientation & 1 & &\\
\hline
Equivalent Diameter & 1 & &\\
\hline
Solidity & 1 & &\\
\hline
Extent & 1 & &\\
\hline
Width/Height & 2 & &\\
\hline
\end{tabular}
\caption{Shape and fixation features used in our model. The final feature dimension for each mask is $33$.} \label{tab:feat}
\end{center}
\end{table}

\textbf{Learning:}  For each dataset, we train a random forest with 30 trees, using a random sampling of $40\%$ of the images. The rest of images are used for testing. The results are averaged on a 10-fold random split of the training and testing set. We use random regression forest to predict the saliency score of an object mask. A random regression forest is an ensemble of decision trees. For each branch node, a feature is selected from a random subset of all features and a decision boundary is set by minimizing the Minimum Square Error (MSE).  The leaf nodes keep the mean value of all training samples that end up in the node. And the final result is a weighted average of all leaf nodes that a testing sample reaches. We choose random forest since our feature vector contains discrete values (Euler Number), which can be easily handled in a decision tree. In the testing phase, each segment is classified independently. We generate the salient object segmentation by averaging the top-$K$ segments at pixel level, similar to~\cite{carreira2010constrained}. We then use simple thresholding to generate the final object masks. As our saliency scores are defined over image segments, this simple strategy leads to accurate object boundaries.

Note that no appearance feature is used in our method, because our goal is to demonstrate the connection between fixation and salient object segmentation.  Our algorithm is independent of the underlying segmentation and fixation prediction algorithms, allowing us to switch between different fixation algorithm or even human fixations.

\subsection{Limits of the model}
Our model contains two separate parts: a segmenter that proposes regions, and a selector that gives each region a saliency score.  In this section we explore the limitation of our model, by replacing each part at a time.  First, we quantify the performance upper-bound of the selector, and then, best achievable results of the segmenter is also presented.

To test the upper-bound of the selector, we train our model on the ground-truth segments of PASCAL-S (e.g.\ Fig.~\ref{fig:objLabelGT}.B) with human fixation maps.  With a perfect segmenter, this model can accurately estimate the saliency of segments using fixation and shape information.  On the test set, it achieves a F-Measure of $0.9201$ with $P = 0.9328$ and $R = 0.7989$.  This result is a strong validation to our motivation, which is to bridge the gap between fixations and salient objects.  It is worth mentioning that this experiment requires a full segmentation of all objects of the entire dataset.  Therefore, PASCAL-S is the only dataset that allows us to test the selector with an ideal segmenter.

Second, we test the upper-bound performance of CPMC segmentation algorithm.  We match the each segment from CPMC to the ground truth object annotations, and greedily choose the segments (out of the first 200 segments) with best overlapping scores.  Again, the result is very positive.  With the first 200 segments, the best CPMC segments achieved an F-Measure of $0.8699$ ($P = 0.8687$, $R = 0.883$) on PASCAL-S dataset.  Similar results are observed in FT (F-Measure = $0.9496$, $P = 0.9494$, $R = 0.9517$) and IS (F-Measure = $0.8416$, $P = 0.8572$, $R = 0.6982$) datasets.

\subsection{Results}
We conduct extensive experiments on three datasets: FT, IS and PASCAL-S using different fixation models. All results are averaged on a 10-fold random split of the training and testing set. The results are grouped into 4 categories:
\begin{description}
\item [Salient Object Algorithms] refer to the 4 algorithms \verb|FT|\cite{achanta2009frequency}, \verb|GC|\cite{cheng2011global}, \verb|PCAS|\cite{margolin2013makes}, and \verb|SF|\cite{perazzi2012saliency} that are originally proposed for salient object segmentation.
\item [CPMC + Fixation Algorithms] refer to the model proposed in our paper.  We choose the top 10 segments for each image, and assign scores based on the fixation map of 7 algorithms: \verb|AIM|\cite{bruce2005saliency}, \verb|AWS|\cite{garcia2012relationship}, \verb|DVA|\cite{hou2008dynamic}, \verb|GBVS|\cite{harel2006graph}, \verb|ITTI|\cite{itti1998model}, \verb|SIG|\cite{hou2012image}, and \verb|SUN|\cite{zhang2008sun}.
\item [Original Fixation Algorithms] refer to the 7 fixation algorithms.  To cancel the effect of center bias, we add a fixed center bias with $\sigma = 0.4$ of the image width, to each generated fixation map.
\item [Baseline Models] refers to 4 other models.  \verb|CPMC Ranking| refers to the original rankings of CPMC, with the same choice of $K = 10$.  \verb|CPMC+Human Fixations| refers to a variation of our model that replaces the algorithm fixation maps with human fixations -supposedly, human map should reflect the saliency of the scene more accurately than algorithms.  \verb|CPMC Best| refers to the salient object maps generated by greedily selecting the best CPMC segments with respect to the ground truth.  This score estimates the upper limit of any algorithm that is based on CPMC segmentation.  Finally \verb|GT Seg + Human Fixations| refers to the method that uses ground-truth segmentations plus human fixations.  This score validates the strong connection between the fixation task and the salient object segmentation task.
\end{description}

\begin{figure*}[htbp]
\centering
\includegraphics[width=\linewidth]{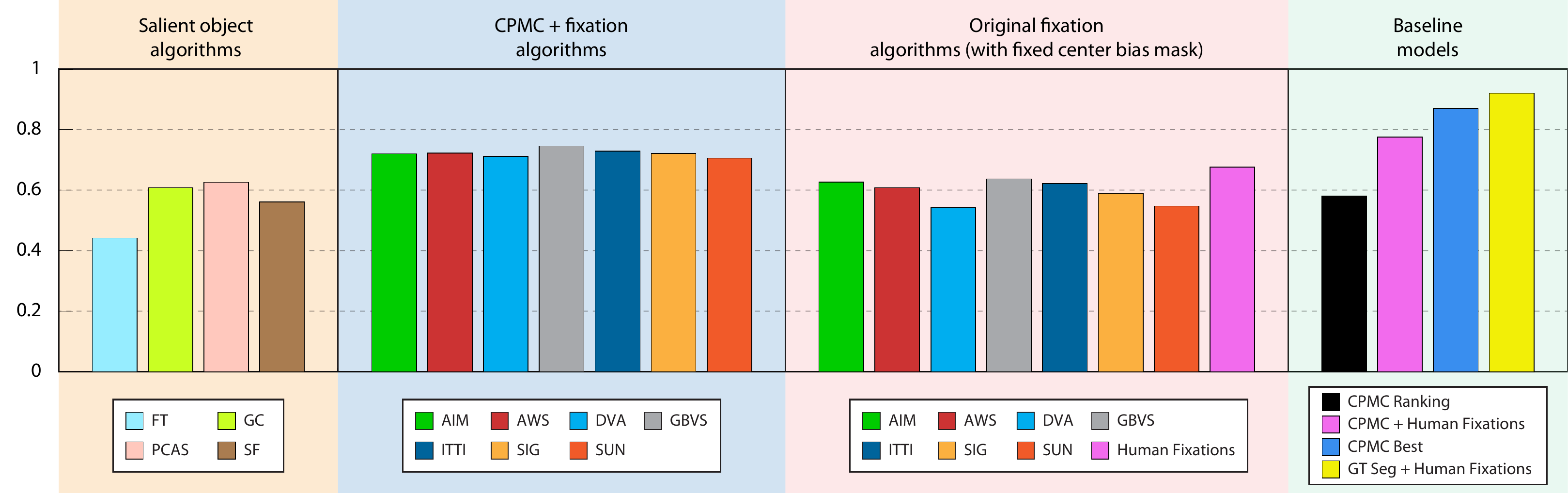}\\
\caption{The F-measures of all algorithms on PASCAL-S dataset.  All CPMC+Fixation results are obtained using top $K=20$ segments.} \label{fig:finalResult}
\end{figure*}

\begin{figure}[htbp]
\centering
\includegraphics[width=0.6\linewidth]{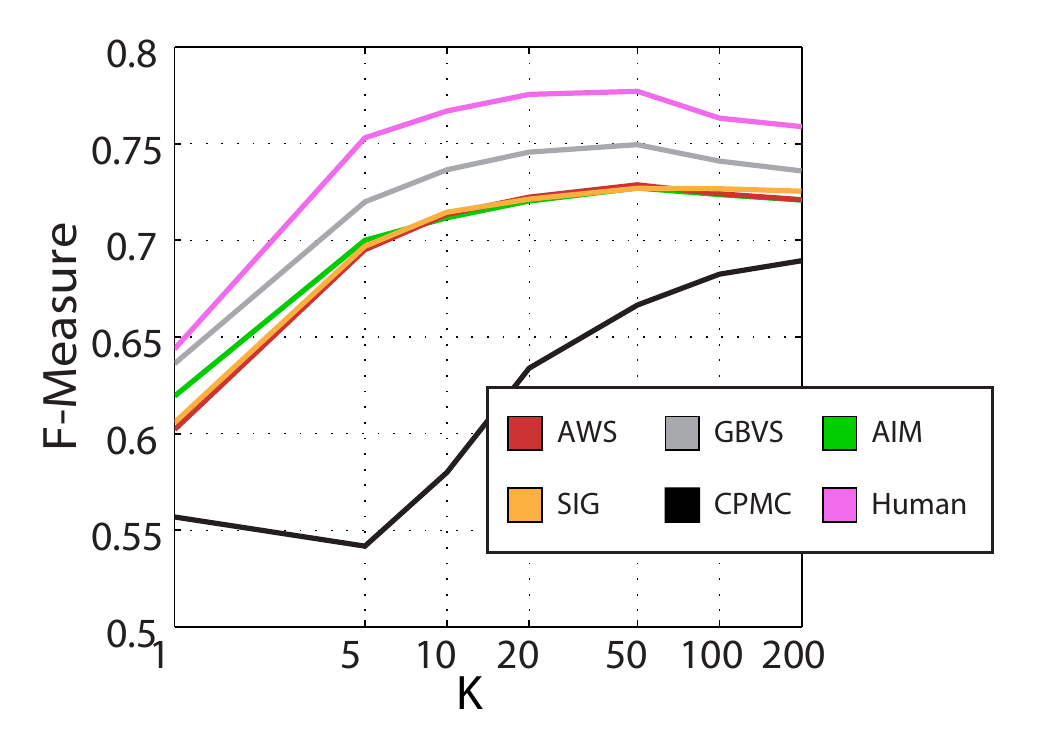}\\
\caption{F-measures of our salient object segmentation method under different choices of $K$, in comparison to CPMC ranking function. Results are reported on the testing set ($60\%$ of the images) of PASCAL-S over 10 random splits. Comparing to the original CPMC ranking function, our method obtains satisfactory F-measures with small $K=20$. } \label{fig:lastFig}
\end{figure}

The full results on PASCAL-S dataset are shown in Fig.~\ref{fig:finalResult} and Fig.~\ref{fig:lastFig}. From these results, we make two key observations. First, our method consistently outperforms the original CPMC ranking function by a large margin, independently of the underlying fixation prediction algorithm. Second, the performance of our model converges much faster than the original CPMC with respect to $K$.  Our model provides decent F-measure with a moderate $K = 20$ segments, while the CPMC ranking function does not converge even with $K=200$\footnote{This might be explained by the diversification ranking used in CPMC.} (see Fig.~\ref{fig:lastFig}).  The finding suggests that our method can effectively segment salient objects using a small number of segments.

The F-measures of each algorithm on all dataset are reported in Tab.~\ref{tab:fmeasures}, using only the first $20$ segments. More results of FT and IS datasets with the combination of 7 fixation algorithms 7 different $K$ can be found in our project website\footnote{\href{http://cbi.gatech.edu/salobj}{http://cbi.gatech.edu/salobj}}. Our methods achieve superior results and consistently outperform all state-of-the-art algorithms on all datasets with a small number of segments. To summarize, we compare our best results ($K=20$) with the best state-of-the-art salient object algorithms in Tab.~\ref{tab:scores}. Our method outperforms the state-of-the-art salient object segmentation algorithms by a large margin. We achieved an improvement of $11.82\%$ , $7.06\%$ and $2.47\%$ on PASCAL-S, IS and FT, respectively, in comparison to the best performing salient object segmentation algorithm. The combination of CPMC+GBVS leads to best results in all \textbf{3} datasets.

\begin{table}
\centering
\begin{tabular} {| c | c | c | c || c | c | c | c | }
\hline
\textbf{Salient Object} & FT & IS & PASCAL-S & \textbf{Baseline Models} & FT & IS & PASCAL-S\\
\hline
FT & 0.7427 & 0.4736 & 0.4325 & CPMC Ranking & 0.2661 & 0.3918 & 0.5799\\
\hline
GC & 0.8383 & 0.6261 & 0.6072 & CPMC + Human & N/A & 0.7863 & 0.7756\\
\hline
PCAS & 0.8646 & \textbf{0.6558} & \textbf{0.6275} & CPMC Best & 0.9496 & 0.8416 & 0.8699\\
\hline
SF & \textbf{0.8850} & 0.5555 & 0.557 & GT Seg. + Human & N/A & N/A & 0.9201\\
\hline
\hline
\textbf{Orig. Fixation} & FT & IS & PASCAL-S & \textbf{CPMC + Fixation} & FT & IS & PASCAL-S\\
\hline
AIM & 0.7148 & 0.4522 & \textbf{0.6216} & AIM & 0.8920 & 0.6728 & 0.7204\\
\hline
AWS & \textbf{0.7240} & \textbf{0.6121} & 0.5906 & AWS & 0.8998 & 0.7241 & 0.7224\\
\hline
DVA & 0.6592 & 0.3764 & 0.5223 & DVA & 0.8700 & 0.6377 & 0.7112\\
\hline
GBVS & 0.7093 & 0.5308 & 0.6186 & GBVS & \textbf{0.9097} & \textbf{0.7264} & \textbf{0.7454}\\
\hline
ITTI & 0.6816 & 0.4452 & 0.6079 & ITTI & 0.8950 & 0.6827 & 0.7288\\
\hline
SIG & 0.6959 & 0.5131 & 0.5850 & SIG & 0.8908 & 0.7255 & 0.7214\\
\hline
SUN & 0.6708 & 0.3314 & 0.5281 & SUN & 0.8635 & 0.6249 & 0.7058\\
\hline
\end{tabular}
\caption{The F-measures of all algorithms on all 3 datasets.} \label{tab:fmeasures}
\end{table}

\begin{table}
\begin{center}
\begin{tabular} {l| c c c}
\hline
\hline
K=20                    & PASCAL-S    & IS     & FT \\
\hline
Fixation Model Used     & GBVS      & GBVS    & GBVS\\
F-Measure               & 0.7457    & 0.7264 & 0.9097\\
Improvements            & \textbf{+11.82}    & \textbf{+7.06}  & \textbf{+2.47} \\
\hline
Best SalObj Model       & PCAS      & PCAS   & SF \\
F-Measure               & 0.6275    & 0.6558 & 0.8850 \\
\hline
\hline
\end{tabular}
\caption{Best results of our model compared to best results of existing salient object algorithms.  Our model achieves better F-measure than all major salient object segmentation algorithms on all \textbf{three} datasets, including the heavily biased FT dataset. Results are reported on the testing set ($60\%$ of the images) over 10 random splits in three datasets.} \label{tab:scores}
\end{center}
\end{table}

\begin{figure*}[p]
\centering
\includegraphics[width=0.65\linewidth]{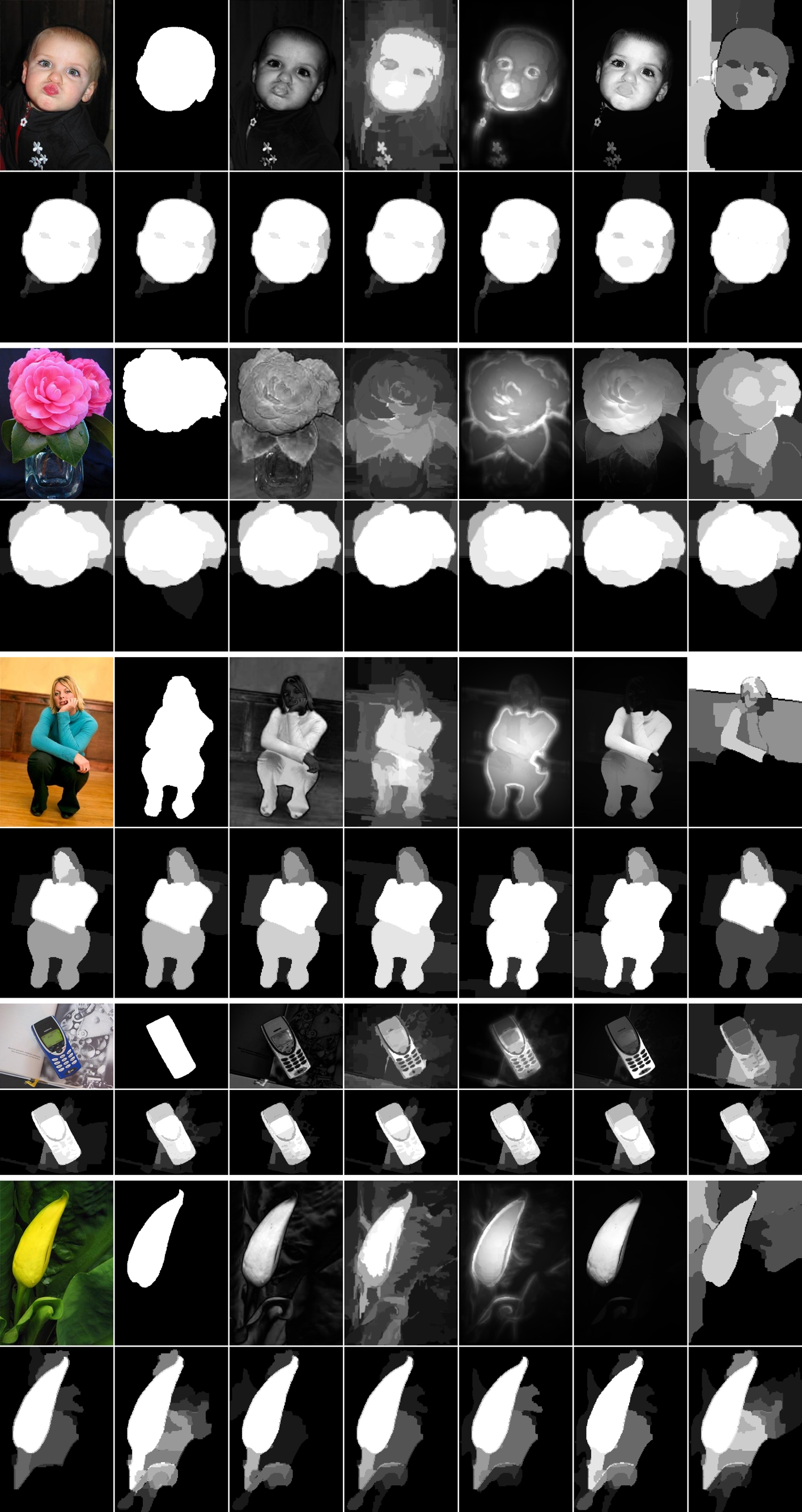}\\
\caption{Visualization of salient object segmentation results on FT. Each two-row compares the results of one image.  The first row includes results from existing methods (Left to Right): Original image, Ground-truth mask, FT, GC, PCAS, SF and CPMC ranking; The second row shows results of our method using different fixations (Left to Right): AIM, AWS, DVA, GBVS, ITTI, SIG and SUN. We are not able to report results using human fixations. The images are selected by sorting the F-measure of our results in a decreasing order. }\label{fig:ftRes}
\end{figure*}

\begin{figure*}[p]
\centering
\includegraphics[width=\linewidth]{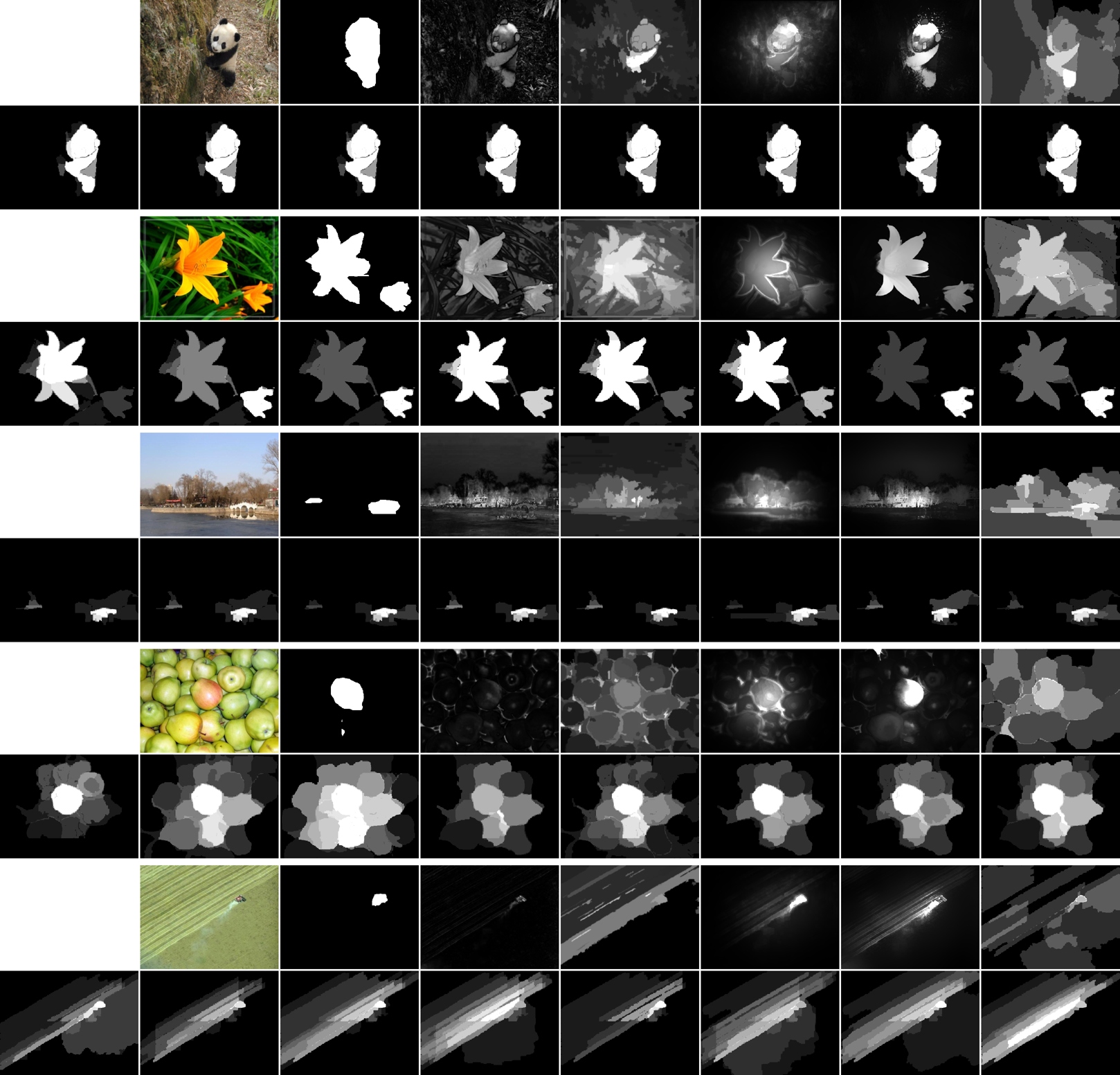}\\
\caption{Visualization of salient object segmentation results on IS. Each two-row compares the results of one image.  The first row includes results from existing methods (Left to Right): Original image, Ground-truth mask, FT, GC, PCAS, SF and CPMC ranking; The second row shows results of our method using different fixations (Left to Right): Human Fixation, AIM, AWS, DVA, GBVS, ITTI, SIG and SUN. The images are selected by sorting the F-measure of our results in a decreasing order. We notice that IS favors sparse saliency maps, since it contains a significant portion of small salient objects.}\label{fig:isRes}
\end{figure*}

\begin{figure*}[p]
\centering
\includegraphics[width=\linewidth]{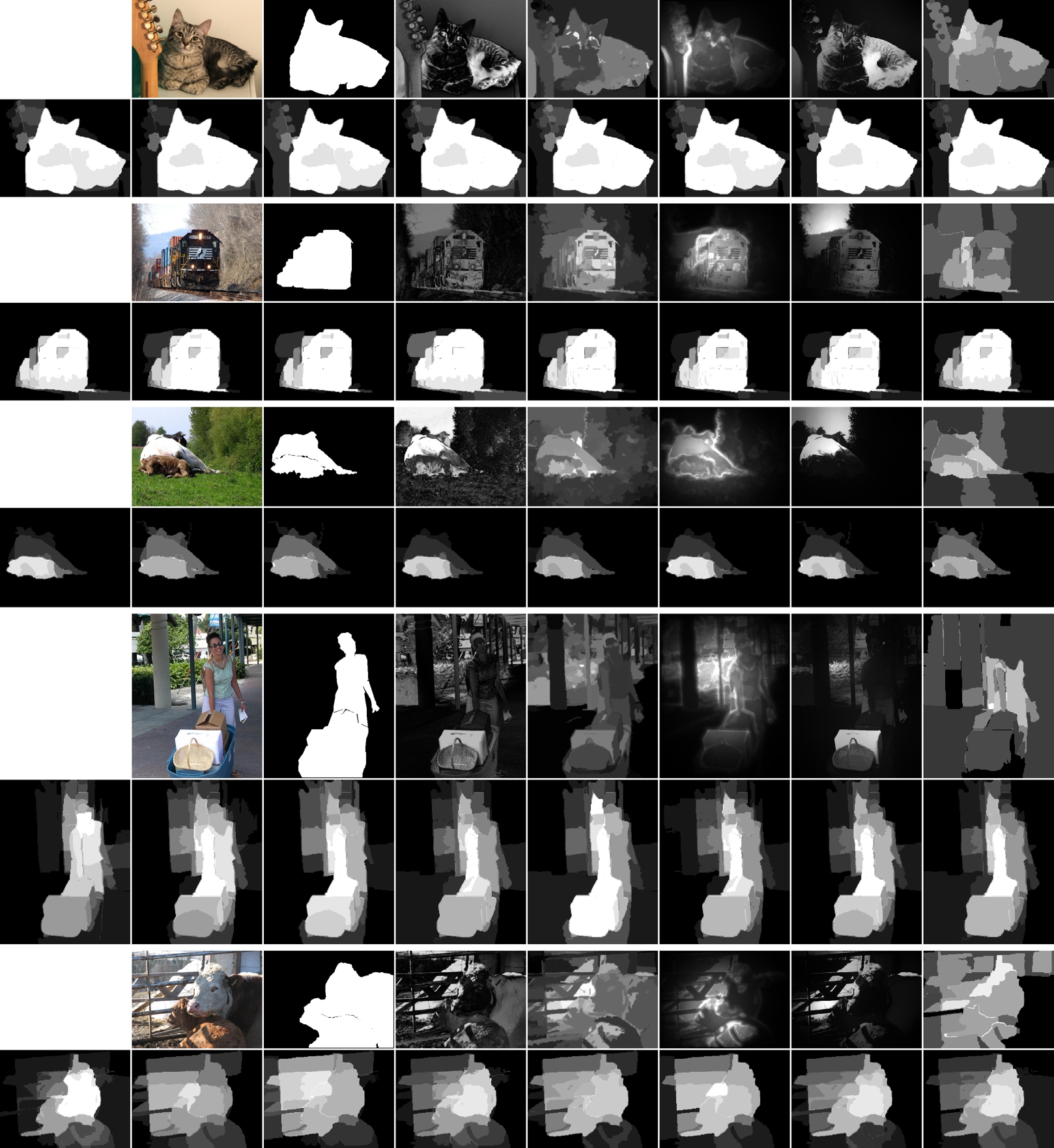}\\
\caption{Visualization of salient object segmentation results on PARSCAL-S. Each two-row compares the results of one image.  The first row includes results from existing methods (Left to Right): Original image, Ground-truth mask, FT, GC, PCAS, SF and CPMC ranking; The second row shows results of our method using different fixations (Left to Right): Human Fixation, AIM, AWS, DVA, GBVS, ITTI, SIG and SUN. The images are selected by sorting the F-measure of our results in a decreasing order.}\label{fig:pascalRes}
\end{figure*}

To better understand our results, we also provide qualitative visualization of salient object masks generated by our methods as well as others. Fig.~\ref{fig:ftRes}-\ref{fig:pascalRes} demonstrate our results using 7 different fixation prediction algorithms in comparison to CPMC ranking function and major salient object segmentation methods, on FT, IS and PARSCAL-S, respectively. We also present our results using human fixations in PASCAL-S and IS.  To illustrate the strength and weakness of our method, we select the results based on the F-measures of each image. For each dataset, the average F-measure of our method decreases from top row to the bottom row. Our method is able to capture the full region of an salient object.  We notice that our method does not segment most of the small salient objects. This is largely due to the output from CPMC using sparse uniform seeds. An object is missing if CPMC does not generate a segment for the object.


\section{Conclusion}
In this paper, we explore the connections between fixation prediction and salient object segmentation by providing a new dataset with both fixations and salient object annotations. We conduct extensive experiments on the dataset for both tasks and compare the results with major benchmarks. Our analysis suggests that the definition of a salient object is highly consistent among human subjects. We also point out significant dataset design bias in major salient object benchmarks. The bias is largely due to deliberately emphasising the concept of saliency. We argue that the problem of salient object segmentation should move beyond the textbook examples of visual saliency. A possible new direction is to look into the strong correlation between fixations and salient objects. Built on top of this connection, we propose a new salient object segmentation algorithm. Our method decouple the problem into a segment generation process, followed a saliency scoring mechanism using fixation prediction. This simple model outperforms state-of-the-art salient object segmentation algorithms on all major datasets. Our dataset together with our method provide a new insight to the challenging problems of both fixation prediction and salient object segmentation.

\clearpage

{
\bibliographystyle{ieee}
\bibliography{egbibShort}
}

\end{document}